 \pdfoutput=1
\documentclass[11pt,a4paper]{article}
\usepackage{acl2015}
\usepackage{times}
\usepackage{url}
\usepackage{latexsym}

\usepackage{multirow}
\usepackage{amsfonts}
\usepackage{amsmath}
\usepackage{graphicx}
\usepackage{eqparbox}

\title{Feature-based Decipherment for Large Vocabulary Machine Translation}

\author{Iftekhar Naim \and Daniel Gildea \\
           Department of Computer Science\\
	    University of Rochester\\
	    Rochester, NY 14627}

\date{}

\begin{document}
\maketitle
\begin{abstract}
Orthographic similarities across languages provide a strong signal for probabilistic decipherment, especially for closely related language pairs.
The existing decipherment models, however, are not well-suited for exploiting these orthographic similarities.
We propose a log-linear model with latent variables that incorporates orthographic similarity features. 
Maximum likelihood training is computationally expensive for the proposed log-linear model. 
To address this challenge, we perform approximate inference via MCMC sampling and contrastive divergence. 
Our results show that the proposed log-linear model with contrastive divergence scales to large vocabularies and outperforms the existing generative decipherment models by exploiting the orthographic features.
\end{abstract}

\section{Introduction}
Word-level translation models are typically learned by applying statistical word alignment algorithms on large bilingual parallel corpora~\cite{Brown:93}. 
However, building a parallel corpus is expensive, and data is limited or even unavailable for many language pairs. 
On the other hand, large monolingual corpora can be easily downloaded from the internet for most languages. 
Decipherment algorithms exploit such monolingual corpora in order to learn translation model parameters, when parallel data is limited or unavailable~\cite{koehn:decipherment:2000,ravi:2011,dou-emnlp14}. 

Existing decipherment methods are predominantly based on probabilistic generative models~\cite{koehn:decipherment:2000,ravi:2011,nuhn-acl14,dou2012large}. 
These models exploit the statistical similarities between the $n$-gram frequencies in the source and the target language, and rely on the Expectation Maximization (EM) algorithm~\cite{Dem77} or its faster approximations.
These existing models, however, do not allow incorporating linguistically motivated features. 
Previous research has shown the effectiveness of incorporating linguistically motivated features for many different unsupervised learning tasks, such as: unsupervised part-of-speech induction~\cite{Berg:10,Haghighi:06}, word alignment~\cite{Ammar:14,Dyer:ACL11}, and grammar induction~\cite{Berg:10}. 
In this paper, we present a feature-rich log-linear model for probabilistic decipherment. 

Words in different languages are often derived from the same source, or borrowed from other languages with minor variations, resulting in substantial phonetic and lexical similarities.
As a result, orthographic features provide crucial information on determining word-level translations for closely related language pairs.
\newcite{haghighi-2008} proposed a generative model for inducing a bilingual lexicon from monolingual text by exploiting orthographic and contextual similarities among the words in two different languages. 
The model proposed by Haghighi et al.\ learns a one-to-one mapping between the words in two languages by analyzing type-level features only, while ignoring the token-level frequencies. 
We propose a decipherment model, that unifies the type-level feature-based approach of Haghighi et al.~with the token-level EM based approaches~\cite{koehn:decipherment:2000,ravi:2011}.

One of the key challenges with the proposed latent variable log-linear models is the high computational complexity of training, as it requires ``normalizing globally'' via summing over all possible observations and latent variables. 
We perform approximate inference using Markov Chain Monte Carlo (MCMC) sampling for scalable training of the log-linear decipherment models. 
The main contributions of this paper are:
\begin{itemize}
\item We propose a feature-based decipherment model that combines both type-level orthographic features and token-level distributional similarities. Our proposed model outperforms the existing EM-based decipherment models.
\item We apply three different MCMC sampling strategies for scalable training and compare them in terms of running time and accuracy. 
Our results show that Contrastive Divergence~\cite{hinton-2002} based MCMC sampling can dramatically improve the speed of the training, while achieving comparable accuracy.
\end{itemize}
\section{Problem Formulation}
Given a source text $\mathcal{F}$ and an independent target corpus $\mathcal{E}$, our goal is to decipher the source text $\mathcal{F}$ by learning the mapping between the words in the source and the target language. 
Although the sentences in the source and target corpus are independent of each other, there exist distributional and lexical similarities among the words of the two languages. 
We aim to automatically learn the translation probabilities $p(f|e)$ by exploiting the similarities between the bigrams in $\mathcal{F}$ and $\mathcal{E}$.

As a simplification step, we break down the sentences in the source and target corpus as a collection of bigrams. 
Let $\mathcal{F}$ contain a collection of source bigrams $f_1 f_2$, and $\mathcal{E}$ contain a collection of target bigrams $e_1 e_2$. 
Let the source and target vocabulary be $V_F$ and $V_E$ respectively. 
Let $N_F$ and $N_E$ be the number of unique bigrams in $\mathcal{F}$ and $\mathcal{E}$ respectively.
We assume that the corpus $\mathcal{F}$ is an encrypted version of a plaintext in the target language.
Each source word $f \in V_F$ is obtained by substituting one of the words $e \in V_E$ in the plaintext. 
However, the mappings between the words in the two languages are unknown, and are learned as latent variables.
\section{Background Research}
\begin{table}
\small
\centering
\begin{tabular}{ | l | l | }
  \hline 
         {Symbol}  & { Meaning} \\  \hline  \hline
        $N_F$  & Number of unique source bigrams \\ \hline
        $N_E$  & Number of unique target bigrams \\ \hline
        	$V_F$ &   Source Vocabulary     \\ \hline 
	$V_E$  &   Target Vocabulary    \\ \hline 
        $V$  & $\max(|V_F|,|V_E|)$  \\ \hline
        $n$ & Number of samples \\ \hline
        $K$ & Beam size for precomputed lists \\ \hline
        $\phi$ & Unigram level feature function \\ \hline
        $\mathbf{\Phi}$  & Bigram level feature function: $\mathbf{\Phi} = \phi_1 + \phi_2$ \\ \hline
\end{tabular}
\caption{Our notations and symbols.}
\label{table:notation}
\end{table}
Existing decipherment models assume that each source bigram $f_1 f_2$ in $\mathcal{F}$ is generated by first generating a target bigram $e_1 e_2$ according to the target language model, and then substituting $e_1$ and $e_2$ with $f_1$ and $f_2$ respectively. 
The generative process is typically modeled via a Hidden Markov Model (HMM) as shown in Figure~\ref{fig:graphical}(a).
The target bigram language model $p(e_1 e_2)$ is trained from the given monolingual target corpus $\mathcal{E}$. 
The translation probabilities $p(f|e)$ are unknown, and learned by maximizing the likelihood of the observed source corpus $\mathcal{F}$:
\begin{eqnarray}
P(\mathcal{F} ) &=& \prod_{f_1 f_2 \in \mathcal{F}} p(f_1 f_2)  \\ 
  &=& \prod_{f_1f_2 \in \mathcal{F}} \sum_{e_1e_2} p(e_1 e_2) p(f_1|e_1) p(f_2|e_2), \nonumber
\end{eqnarray}
where $e_1$ and $e_2$ are the latent variables, indicating the target words in $V_E$ corresponding to $f_1$ and $f_2$ respectively.
The log-likelihood function with latent variables is non-convex, and several methods have been proposed for maximizing it. 
\subsection{Expectation-Maximization (EM)}
The Expectation-Maximization (EM)~\cite{Dem77} algorithm has been widely applied for solving the decipherment problem~\cite{Knight-decipher99,koehn:decipherment:2000}. 
In the E-step, for each source bigram $f_1 f_2$, we estimate the expected counts of the latent variables $e_1$ and $e_2$ over all the target words in $V_E$. 
In the M-step, the expected counts are normalized to obtain the translation probabilities $p(f|e)$. 
The computational complexity of the EM algorithm is $O(N_F V^2)$ and the memory complexity is $O(V^2)$, where $N_F$ is the number of unique bigrams in $\mathcal{F}$ and $V = \max(|V_F|, |V_E|)$.  
As a result, the regular EM algorithm is prohibitively expensive for large vocabulary sizes, both in terms of running time and memory consumption.

To address this challenge, Ravi and Knight~\shortcite{ravi:2011} proposed the Iterative EM algorithm, which starts with the $K$ most frequent words from $\mathcal{F}$  and $\mathcal{E}$ and performs EM-based decipherment. 
Next, the source and target vocabularies are iteratively extended by $K$ new words, while pruning low probability entries from the probability table. 
The computational complexity of each iteration becomes $O(N_F K^2)$.
\subsection{Bayesian Decipherment using Gibbs Sampling}
Ravi and Knight~\shortcite{ravi:2011} proposed a Gibbs sampling based Bayesian Decipherment strategy. 
For each observed source bigram $f_1 f_2$, the Gibbs sampling approach starts with an initial target bigram $e_1 e_2$, and alternately fixes one of the target words and replaces the other with a randomly chosen sample. 
When $e_1$ is fixed, a new sample $e_2^{new}$ is drawn from the probability distribution $p(e_1 e_2^{new}) p(f_2|e_2^{new})$. 
Next, we fix $e_2$ and sample $e_1^{new}$, and continue alternating until $n$ samples are collected. 
Bayesian decipherment reduces memory consumption via Gibbs sampling. 
The probability table remains sparse, since only a small number of word pairs $(f,e)$ will be observed together in the samples. 
\subsection{Slice Sampling}
To draw each sample via Gibbs sampling, we need to estimate the probabilities of choosing each target word $e \in V_E$, which requires $O(V)$ operations. 
To address this issue, Dou et al.~\shortcite{dou2012large} proposed a slice sampling approach with precomputed top-$K$ lists. 
Similar to Gibbs sampling, for each source bigram $f_1 f_2$, the slice sampling approach starts with one initial target bigram $e_1 e_2$, and alternately replaces either $e_1$ or $e_2$ while keeping the other one fixed.
In order to replace $e_1$ with a new sample $e_1^{new}$, we sample a random threshold $T$ uniformly between 0 and $p(e_1 e_2) p(f_1|e_1)$.
Next, we uniformly sample an $e_1^{new}$ from all the candidates $e_1^\prime$ such that $p(e_1^\prime e_2) p(f_1|e_1^\prime) > T$. 
While sampling $T$ is straightforward, the second sampling stage requires finding all the candidates, which again takes $O(V)$ computation.
Dou et al.~\shortcite{dou2012large} addressed this challenge by precomputing sorted top-$K$ word lists for both $p(f|e)$ and $p(e_1, e_2)$.
While sampling $e_1$, it tries to generate all the candidates by looking only at the top-$K$ lists for $p(e_1^\prime| f_1)$ and the top $K$ list for  $p(e_1^\prime e_2)$. 
Even though slice sampling with top-$K$ lists is faster than Gibbs sampling on average, sometimes the top-$K$ lists fail to provide all the candidates, and it needs to fall back to sampling from the entire vocabulary, which requires $O(V)$ operations.
\subsection{Beam Search}
Nuhn et al.~\shortcite{nuhn-acl13,nuhn-acl14} showed that Beam Search can significantly improve the speed of EM-based decipherment, while providing comparable or even slightly better accuracy. 
Beam search prunes less promising latent states by maintaining two constant-sized beams, one for the translation probabilities $p(f|e)$ and one for the target bigram probabilities $p(e_1 e_2)$ -- reducing the computational complexity to $O(N_F)$. 
Furthermore, it saves memory because many of the word pairs $(f,e)$ are never considered due to not being in the beam. 
\subsection{Feature-based Generative Models}
Feature-based representations have previously been explored under the generative setting. 
Haghighi et al.~\shortcite{haghighi-2008} proposed a Canonical Correlation Analysis (CCA) based model for automatically learning the mapping between the words in two languages from monolingual corpora only. 
They exploited the orthographic and contextual features between the word types, but ignored the token-level frequencies. 
Ravi~\shortcite{ravi-2013} proposed a Bayesian decipherment model based on hash sampling, which takes advantage of feature-based similarities between source and target words. 
However, the feature representation was not integrated with their decipherment model, and was only used for efficiently sampling candidate target translations for each source word. 
Furthermore, the feature based hash sampling included only contextual features, and did not consider orthographic features. 
In contrast, our log-linear model integrates both type-level orthographic features and token-level bigram frequencies.
\section{Feature-based Decipherment}
\begin{figure}[tb]
\centering
\includegraphics[width=0.45\textwidth]{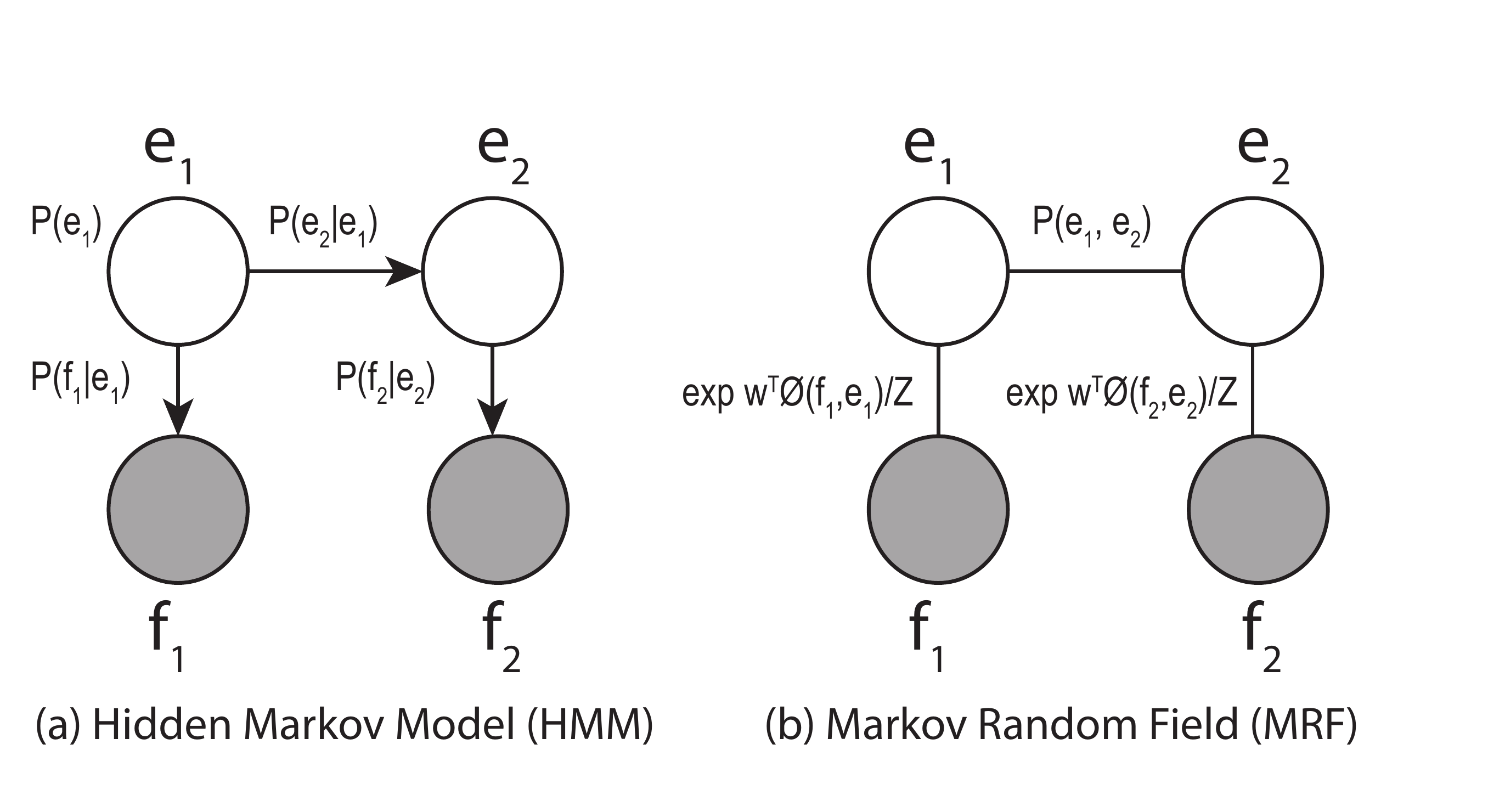}
\caption{The graphical models for the existing directed HMM and the proposed undirected MRF.}
\label{fig:graphical}
\end{figure}

Our feature-based decipherment model is based on a chain structured Markov Random Field (Figure~\ref{fig:graphical}(b)), which jointly models the observed source bigrams $f_1f_2$ and corresponding latent target bigram $e_1 e_2$. 
For each source word $f \in V_F$, we have a latent variable $e \in V_E$ indicating the corresponding target word.
The joint probability distribution:
\begin{equation}
p(f_1 f_2, e_1 e_2) = \frac{1}{Z_{\mathbf{w}}} \exp{\mathbf{w}^T\mathbf{\Phi}(f_1f_2, e_1e_2)} p(e_1e_2),
\end{equation}
where $\mathbf{\Phi}(f_1f_2, e_1e_2)$ is the feature function for the given source and the target bigrams, $\mathbf{w}$ is the model parameters, and $Z_{\mathbf{w}}$ is the normalization term. We assume that the bigram feature function decomposes linearly over the two unigrams:
\begin{equation}
\mathbf{\Phi}(f_1f_2, e_1e_2) = \phi(f_1, e_1) + \phi(f_2, e_2)
\end{equation}
The normalization term is:
\begin{equation*}
Z_{\mathbf{w}} = \sum_{f_1f_2} \sum_{e_1 e_2} p(e_1e_2) \exp{\mathbf{w}^T\mathbf{\Phi}(f_1f_2, e_1e_2)}
\end{equation*}
The gradient of the joint log-likelihood is:
\begin{align*}
 \label{eq:grad}
\frac{\partial L} { \partial \mathbf{w}} &=    \mathbb{E}_{e_1 e_2|f_1f_2} \left[ \mathbf{\Phi}(f_1 f_2, e_1 e_2) \right  ] - \\
   &\qquad \mathbb{E}_{f_1f_2,e_1e_2} \left[ \mathbf{\Phi}(f_1 f_2, e_1 e_2) \right  ] \\
&=  \mathbb{E}^{Forced} - \mathbb{E}^{Full}
\end{align*}
Here, the first term is the expectation with respect to the empirical data distribution. 
We refer to it as the ``Forced Expectation", as the source text is assumed to be given. 
The second term is the expectation with respect to our model distribution, and referred to as ``Full Expectation".
In theory, we can apply gradient descent or other off-the-shelf optimization techniques to optimize the conditional log-likelihood. 
However, exact estimation of the gradient is computationally expensive, as discussed in the next sub-sections.

\subsection{Estimating Forced Expectation ($\mathbb{E}^{Forced} $)}
We estimate the forced expectation over latent variables using the following equation:
\begin{multline}
\mathbb{E}^{Forced} =  \sum_{f_1f_2 \in \mathcal{F}} \frac{1}{Z(f_1f_2)}  \sum_{e_1e_2 \in V_E^2}  \biggl [ p(e_1 e_2)\\  \exp{\mathbf{w}^T\mathbf{\Phi}(f_1f_2, e_1e_2)} \biggr ] \mathbf{\Phi}(f_1 f_2, e_1 e_2),
\end{multline}
where $Z(f_1f_2)$ is the normalization term given $f_1 f_2$: 
\begin{equation*}
Z(f_1f_2) = \sum_{e_1e_2 \in V_E^2} p(e_1 e_2)  \exp{\mathbf{w}^T\mathbf{\Phi}(f_1f_2, e_1e_2)}.
\end{equation*}
\normalsize
For each observed $f_1 f_2 \in \mathcal{F}$, we need to sum over all possible $e_1 e_2 \in V_E^2$, which requires $O(N_F V^2 )$ computation. 
\subsection{Estimating Full Expectation ($\mathbb{E}^{Full} $)}
For the full expectation, we assume that both the source text and latent variables are unknown.
We estimate it by summing over all the possible source bigrams $f_1f_2$, and associated latent variables $e_1e_2$:
\begin{multline}
\mathbb{E}^{Full} =  \frac{1}{Z_g}   \sum_{f_1 f_2 \in V_F^2} \sum_{e_1 e_2 \in V_E^2}  \biggl [ p(e_1 e_2) \\ \exp{\mathbf{w}^T\mathbf{\Phi}(f_1f_2, e_1e_2)} \biggr ] \mathbf{\Phi}(f_1 f_2, e_1 e_2),
\end{multline}
where $Z_g$ is the global normalization term:
\begin{multline*}
Z_g = \sum_{f_1 f_2 \in V_F^2} \sum_{e_1e_2 \in V_E^2} p(e_1 e_2) \\ \exp{\mathbf{w}^T\mathbf{\Phi}(f_1f_2, e_1e_2)}.
\end{multline*}
\normalsize
The computational complexity is $O(V^4)$.
\begin{table}
\small
\begin{tabular}{ || l || l || }
  \hline 
         {Method}  & { Complexity per Iteration} \\  \hline  \hline
	EM &   $O(N_F V^2)$     \\ \hline 
	EM + Slice &   $O(N_F nV)$, but often faster    \\ \hline 
        Log-linear Exact  & $O(N_F V^2 + V^4)$   \\ \hline
        Log-linear + Gibbs  & $O(N_F V n + V n^2)$  \\ \hline
        Log-linear + IMH + Gibbs  & $O(N_F n + V n^2)$ \\ \hline
        Log-linear + CD  & $O(N_F n)$ \\ \hline
    \hline
\end{tabular}
\caption{The worst case computational complexities for different decipherment algorithms}
\label{table:complexity}
\end{table}

\section{MCMC Sampling for Faster Training} 
The overall computational complexity of estimating the exact gradient is $O(N_F V^2 + V^4)$, which is infeasible for decipherment even with a modest-sized vocabulary. 
Instead, we apply several different MCMC sampling methods to approximately estimate the forced and full expectations.

\subsection{Gibbs Sampling}
\subsubsection{Gibbs Sampling for Approximating Forced Expectation}
Instead of summing over all target bigrams $e_1 e_2$, we approximate the forced expectation by taking $n$ samples of $e_1 e_2$ for each observed $f_1 f_2$, and take an average of the features for these samples. 
For each observed $f_1 f_2$, the following steps are taken:
\begin{itemize}
\item Start with an initial target bigram $e_1 e_2$.
\item Fix $e_2$ and sample $e_1^{new}$ according to the following probability distribution:
\begin{multline*}
P(e_1^{new}  | e_2, f_1f_2) = \frac{1}{Z_{gibbs}} \biggl [ p(e_1^{new} e_2) \\  \exp{\mathbf{w}^T\mathbf{\Phi}(f_1f_2, e_1^{new} e_2)} \biggr ]
\end{multline*}
where
\begin{equation*}
Z_{gibbs} = \sum_{e_1} p(e_1 e_2)  \exp{\mathbf{w}^T\mathbf{\Phi}(f_1f_2, e_1e_2)}
\end{equation*}

\item Next, fix $e_1$ and draw a new sample $e_2$ similarly according to $P(e_2^{new}  | e_1, f_1f_2)$, and continue sampling $e_1$ and $e_2$ alternately until $n$ samples are drawn.
\end{itemize}
Drawing each sample requires $O(V)$ operations, as we need to estimate the normalization term $Z_{gibbs}$. 
The computational complexity of estimating the forced expectation becomes: $O(N_F V n)$, which is expensive as $V$ can be large. 
\subsubsection{Gibbs Sampling for Approximating Full Expectation}
To efficiently estimate the full expectation, we sample $n$ source bigrams $f_1 f_2$ from our model. The Gibbs sampling procedure is:
\begin{itemize}
\item Start with an initial random $f_1 f_2$.

\item Fix $f_2$, and sample a new $f_1$ according to $p(f_1 | f_2)$:
\begin{multline*}
p(f_1|f_2) = \frac{1}{Z_{gibbs}^\prime} \sum_{e_1} \sum_{e_2} \biggl [  p(e_1 e_2) \\  \exp{\mathbf{w}^T\mathbf{\Phi}(f_1f_2, e_1e_2)} \biggr ]
\end{multline*}
where 
\begin{multline*}
Z_{gibbs}^\prime = \sum_{f_1} \sum_{e_1} \sum_{e_2} \biggl [ p(e_1 e_2) \\ \exp{\mathbf{w}^T\mathbf{\Phi}(f_1f_2, e_1e_2)} \biggr ]
\end{multline*}
\item Next fix $f_1$ and sample $f_2$ according to $P(f_2|f_1)$. Continue alternating until $n$ samples are drawn.
\end{itemize}
The computational complexity of exactly estimating $p(f_1|f_2)$ is $O(V^3)$, resulting in the computational complexity $O(V^3 n)$, which is infeasible. 
However, instead of summing over all possible $e_1 e_2$, we can approximate via sampling. 
For each $f_1f_2$, we first sample $n$ samples $e_1 e_2$ according to $p(e_1 e_2)$. 
Let $S$ be the set of $n$ samples of target bigrams.
Next, we approximate $p(f_1|f_2)$ as:
\begin{equation*}
p(f_1|f_2) = \frac{1}{Z_{approx}} \sum_{e_1 e_2 \in S} \exp{\mathbf{w}^T\mathbf{\Phi}(f_1f_2, e_1e_2)}
\end{equation*}
where 
\begin{equation*}
Z_{approx} = \sum_{f_1} \sum_{e_1e_2 \in S} \exp{\mathbf{w}^T\mathbf{\Phi}(f_1f_2, e_1e_2)}
\end{equation*}
This reduces the computational complexity to $O(Vn^2)$. 
\subsection{Independent Metropolis Hastings (IMH)}
The Gibbs sampling for our log-linear model is slow as it requires normalizing the sampling probabilities over the entire vocabulary. 
To address this challenge, we apply Independent Metropolis Hastings (IMH) sampling, which relies on a proposal distribution and does not require normalization. 
However, finding an appropriate proposal distribution can sometimes be challenging, as it needs to be close to the true distribution for faster mixing and must be easy to sample from.

For the forced expectation, one possibility is to use the bigram language model $p(e_1 e_2)$ as a proposal distribution.
However, the bigram language model did not work well in practice. 
Since $p(e_1 e_2)$ does not depend on $f_1 f_2$, it resulted in slow mixing and exhibited a bias towards highly frequent target words.

Instead, we chose an approximation of $p(e_1 e_2|f_1 f_2) $ as our proposal distribution. 
To simplify sampling, we assume $e_1$ and $e_2$ to be independent of each other for any given $f_1 f_2$.
Therefore, the proposal distribution $q(e_1 e_2 | f_1 f_2) = q_u(e_1|f_1) q_u(e_2|f_2)$, where $q_u(e|f)$ is a probability distribution over target unigrams for a given source unigram.
We define $q_u(e|f)$ as follows:
\begin{equation*}
q_u( e | f) = (1-p_b) q_s(f | e) + p_b \frac{1}{V}
\end{equation*}
where $p_b$ is a small back-off probability with which we fall back to the uniform distribution over target unigrams. The other term $q_s(e|f)$ is a distribution over the target words $e$ for which $(f,e) \in \mathbf{w}$:
\[
    q_s(e|f)= 
\begin{cases}
     \frac{1}{Z_{imh}} \exp{\mathbf{w}^T \phi(f, e)} ,& \text{if } (f,e) \in \mathbf{w}\\
    0,              & \text{otherwise}.
\end{cases}
\]
Here, $Z_{imh}$ is a normalization term over all the $e$ such that $(f,e) \in \mathbf{w}$.
The weight vector $\mathbf{w}$ is sparse, as only a small number of translation features $(f,e)$ (Section~\ref{sec:feature}) are observed during sampling. 
Furthermore, we update $q_s$ only once every 5 iterations of gradient descent.

The actual target distribution is: 
\begin{equation}
p (e_1e_2|f_1f_2) \propto p(e_1 e_2) \exp{\mathbf{w}^T\mathbf{\Phi}(f_1f_2, e_1e_2)} 
\end{equation}

For each $f_1 f_2  \in \mathcal{F}$, we take the following steps during sampling:
\begin{itemize}
\item Start with an initial English bigram: $\langle e_1 e_2\rangle^{0}$
\item Let the current sample be $\langle e_1 e_2 \rangle^{i}$. Next, sample ${\langle e_1 e_2 \rangle}^{i+1}$ from the proposal distribution $q(e_1 e_2|f_1 f_2)$.
\item Accept the new sample with the probability:
\begin{equation*}
P_a = \frac{p( \langle e_1 e_2 \rangle^{i+1}|f_1f_2)} {p(\langle e_1 e_2 \rangle^{i}|f_1f_2) } \frac{q(\langle e_1e_2 \rangle^{i}|f_1f_2)}{q(\langle e_1e_2 \rangle^{i+1}|f_1f_2)}
\end{equation*}
\end {itemize}
The IMH sampling reduces the complexity of the forced expectation estimation to $O(N_F n)$~\footnote{Ignoring the cost of estimating $q_s(e|f)$, which occurs only once every 5 iterations.}, which is significantly less than the complexity of $O(N_F V n )$ in the case of Gibbs sampling.
However, we could not apply IMH while estimating the full expectation, as finding a suitable proposal distribution is more complicated. Therefore, the overall complexity remains: $O(N_F n + Vn^2)$.

 
\subsection{Contrastive Divergence Based Sampling}
The main reason for the slow training of the proposed log-linear model is the high computational cost of estimating the partition function $Z_g$ of our MRF model when estimating the full expectation. 
A similar problem arises while training deep neural networks. 
An increasingly popular technique to address this issue is to perform Contrastive Divergence~\cite{hinton-2002}, which allows us to avoid estimating the partition function.

For each observed source bigram $f_1 f_2 \in \mathcal{F}$, the contrastive divergence sampling procedure works as follows:

\begin{itemize}
\item Sample a target bigram $e_1 e_2$ according to the distribution $p(e_1 e_2| f_1 f_2)$. We perform this step using Independent Metropolis Hastings, as discussed in the previous section.
\item Sample a reconstructed source bigram $ \langle f_1 f_2 \rangle^{recon}$ by sampling from the distribution $p(f_1 f_2 | e_1 e_2)$, again via Independent Metropolis Hastings. 
\end{itemize}
We take $n$ such samples of $e_1 e_2$ and corresponding $ \langle f_1 f_2  \rangle^{recon}$. 
For each sample and reconstruction pair, we update the weight vector by an approximation of the gradient:
\begin{equation*}
\frac{\partial L} { \partial \mathbf{w}} \approx  \mathbf{\Phi}(\langle f_1 f_2\rangle^{data}, e_1 e_2) -   \mathbf{\Phi}(\langle f_1 f_2\rangle ^{recon}, e_1 e_2)
 \end{equation*}
\section{Feature Design}
\label{sec:feature}
We included the following unigram-level features:
\begin{itemize}
\item \emph{Translation Features:} each $(f, e)$ word pair, where $f  \in V_F $ and $e \in V_E$, is a potential feature in our model. While there are $O(V^2)$ such possible features, we only include the ones that are observed during sampling. 
Therefore, our feature weights $\mathbf{w}$ is a sparse vector, with most of the entries zero.
\item \emph{Orthographic Features:} we incorporated an orthographic feature based on the normalized edit-distance. 
For a word pair $(e, f)$, the orthographic feature is triggered if the normalized edit distance is less than a threshold (set to 0.3 in our experiments).
\end{itemize}
The set of features can further be extended by including context window based features~\cite{haghighi-2008,ravi-2013} and topic features.

\section{Experiments and Results}

\subsection{Datasets}
We experimented with two closely related language pairs: (1) Spanish and English and (2) French and English. 
For Spanish/English, we experimented with a subset of the OPUS Subtitle corpus~\cite{Tiedemann:RANLP5}. 
For French/English,  we used the Hansard corpus~\cite{brown-acl91}, containing parallel French and English text from the proceedings of the Canadian Parliament. 
In order to have a non-parallel setup, we extracted monolingual text from different sections of the French and English text.
The detailed description of the two datasets are provided below:
\begin{table}
\small
\begin{tabular}{ | l | l | l | l |}
  \hline 
         {Dataset}  & {Num. Sentences} & {$|V_E|$} & {$|V_F|$} \\  \hline \hline
	OPUS   & $19.77K$ (1128 unique)  & 579  & 411 \\ \hline 
	Hansard-100 & 100  &  358  & 371 \\ \hline 
        Hansard-1000 & 1000 & 2957 & 3082 \\ \hline 
\end{tabular}
\caption{Statistics on the datasets used in our experiments.}
\label{table:datasets}
\end{table}

\textbf{OPUS Subtitle Dataset:} the OPUS dataset is a smaller pre-processed subset of the original larger OPUS Spanish/English parallel corpora.
The dataset consists of short sentences in Spanish and English, each of which is a movie subtitle. 
The same dataset has been used in several previous decipherment experiments~\cite{ravi:2011,nuhn-acl14,ravi-2013}.

\textbf{Hansard Dataset:}
The Hansard dataset contains parallel text from the Canadian Parliament Proceedings. 
We experimented with two datasets:
\begin{itemize}
\item {\textbf{Hansard-100:}} The French text consists of the first 100 sentences and the English text consists of the second 100 sentences.
\item {\textbf{Hansard-1000:}} The French text consists of the first 1000 sentences and the English text consists of the second 1000 sentences.
\end{itemize}

Table~\ref{table:datasets} provides some statistics on the three datasets used in our experiments.
Due to the relatively small vocabulary size of OPUS and Hansard-100 dataset, we were able to run all 4 versions of the log-linear model and compare with the exact EM-based decipherment.
The Hansard-1000 dataset, however, is too large to run the exact EM and some of the inexact log-linear models (e.g., Gibbs sampling and IMH + Gibbs). 
As a result, we only applied the fastest log-linear model with contrastive divergence on the Hansard-1000 dataset.

\subsection{Evaluation}
We evaluate the accuracy of decipherment by the percentage of source words that are mapped to the correct target translation.
The correct translation for each source word was determined automatically using the Google Translation API. 
While the Google Translation API did a fair job of translating the French and Spanish words to English, it returned only a single target translation. 
We noticed occasional cases where the decipherment algorithm retrieved the correct translation, but it did not get the credit because of not matching with the translation from the API. 

Additionally, we performed Viterbi decoding on the sentences in a small held-out test corpus from the OPUS dataset, and compared the BLEU scores with the previously published results on the same training and test sets~\cite{ravi:2011,nuhn-acl14,ravi-2013}.
\subsection{Results}
\begin{table*}
\begin{center}
\footnotesize
\begin{tabular}{ || l || r | r || r | r || r | r ||}
  \hline 
  \multirow{2}{*}{Method} & \multicolumn{2}{ c ||}{OPUS} & \multicolumn{2}{ c ||}{Hansard-100} & \multicolumn{2}{ c ||}{Hansard-1000} \\ \cline{2-7}
     &  Time & Acc ($\%$) & Time & Acc ($\%$) & Time & Acc ($\%$)  \\  
    \hline
	EM                 & 520.2s & 6.04                     & 188.0s & 2.96                   & -- & --      \\ \hline 
        Log-linear + Gibbs         & 429.7s & \textbf{8.63}            & 207.3s & \textbf{14.02}         & -- & --      \\ \hline
        Log-linear + IMH + Gibbs   & 61.6s & \textbf{8.46}             & 39.0s &  \textbf{13.21}         & -- & --      \\ \hline
        Log-linear + CD            & \textbf{15.1s} & \textbf{8.46}    & \textbf{7.77s} & \textbf{12.93} & \textbf{401.0s} & \textbf{15.08} \\ \hline
        Log-linear + CD (No ortho) & \textbf{15.3s} & 1.89             & \textbf{7.70s} & 3.50           & \textbf{396.4s}  & 2.66 \\ \hline
    \hline
\end{tabular}
\end{center}
\caption{The running time per iteration and accuracy of decipherment.}
\label{table:result}
\end{table*}
We experimented with three versions of our log-linear decipherment algorithms: (1) Gibbs Sampling, (2) IMH and Gibbs Sampling, and (3) Contrastive Divergence (CD). 
To determine the impact of the orthographic features, the Contrastive Divergence based log-linear model was tested both with and without the orthographic features. 
We compared the log-linear models with the exact EM algorithm~\cite{koehn:decipherment:2000,ravi:2011}. 
We could not include the exact log-linear model in our experiments due to the extremely slow training. 
The number of iterations was fixed to 50 for all five methods. 
For the sampling based methods, we set the number of samples $n = 50$.
\begin{table}
\small
\begin{tabular}{ | l | r | }
  \hline 
        {Method}  & {BLEU (\%)} \\  \hline \hline
	EM  (Ravi and Knight, 2011) & 15.3 \\
	EM + Beam Search (Nuhn and Ney, 2014) & 15.7  \\ \hline
        Log-linear + Gibbs  & 18.9 \\
        Log-linear + IMH  & 18.8 \\
        Log-linear + CD  & 18.8 \\ \hline 
\end{tabular}
\caption{Comparison of MT performance on the OPUS dataset using bigram language model.}
\label{table:bleu}
\end{table}

For the log-linear model with no orthographic features, we initialized all the feature weights to zero.  
We do not store these initial weights in memory, as they are all set to zero by default.
When we included the orthographic features, we initialized the weight of the orthographic match feature to 1.0 to encourage translation pairs with high orthographic similarity. 
Furthermore, for each word pair $(f,e)$ with high orthographic similarity, we assigned a small positive weight (0.1).
This initialization allowed the proposal distribution to sample orthographically similar target words for each source word.
For the exact EM, we initialized the translation probabilities uniformly and stored the entire probability table. 

We applied all four log-linear models and the exact EM on the OPUS and the Hansard-100 datasets. 
On the Hansard-1000 dataset, we could only apply the Contrastive Divergence based log-linear model (with and without orthographic features) due to its large vocabulary sizes. 
Table~\ref{table:result} reports the accuracy and the running time per iteration for all the methods on the three datasets. 
The BLEU scores for the OPUS dataset are reported in Table~\ref{table:bleu}.
A bigram language model was used for all the models.
Table~\ref{table:example} shows a few examples for which the log-linear model performed better due to orthographic features.
\begin{table}[b]
\small
\begin{tabular}{ | l l || l l |}
  \hline 
        \multicolumn{2}{| c ||}{OPUS} & \multicolumn{2}{ c |}{Hansard-1000} \\  \hline
        \emph{Spanish}  & \emph{English} & \emph{French} & \emph{English} \\  \hline 
	excelente & excellent & criminel & criminal \\ 
	minuto & minute & particulier & particular \\  
	silencio & silence & sociaux & social  \\ \ 
	perfecto & perfect & secteur & sector \\  
    \hline
\end{tabular}
\caption{A few sample examples, for which orthographic features helped.}
\label{table:example}
\end{table}
\section{Discussion and Future Work}
We notice that all the log-linear models with orthographic features outperformed the EM-based methods. 
The only log-linear model which performed much worse was the one which lacked the orthographic features.
This result emphasizes the importance of orthographic features for decipherment between closely related language pairs.
The margin of improvement due to orthographic features was bigger for the Hansard datasets than that for the OPUS dataset. 
It is expected, as the lexical similarity between French and English is higher than that for Spanish and English.   
The Contrastive Divergence based log-linear model achieved comparable accuracy to the two other log-linear models, despite being orders of magnitude faster. 
Furthermore, the log-linear models resulted in better translations, as they obtained significantly higher BLEU score on the OPUS dataset (Table~\ref{table:bleu}).

While the orthographic features provide huge improvements in decipherment accuracy, they also introduce new errors. 
For example, the Spanish word ``madre" means ``mother" in English, but our model gave highest score to the English word ``made" due to the high orthographic similarity. 
However, such error cases are negligible compared to the improvement.

In this paper, we assumed no parallel data is available, and experimented with fairly simple initialization strategies. 
However, the objective functions for both EM and the latent variable log-linear model are non-convex, and the results may vary drastically based on initialization~\cite{Berg-Kirkpatrick-2013}.
In future, we would like to start with a small parallel corpora, and initialize the decipherment models with the parameters learned from the small parallel corpora~\cite{dou-emnlp14}.
We would also like to experiment with a more sophisticated translation model that incorporates NULL words, local reordering of neighboring words, and word fertilities~\cite{ravi-2013}. 
Finally, we would like to incorporate more flexible non-local features, which are not supported by the feature-based directed graphical models, such as Feature-HMM~\cite{Berg:10}. 

\section{Conclusion}
We presented a feature-based decipherment system using latent variable log-linear models.
The proposed models take advantage of the orthographic similarities between closely related languages, and outperform the existing EM-based models. 
The Contrastive Divergence based variant provided the best trade-off between speed and accuracy.

\bibliographystyle{acl}
\bibliography{extra}

\end{document}